\documentclass{article}

\usepackage{arxiv}

\usepackage[utf8]{inputenc} 
\usepackage[T1]{fontenc}    
\usepackage{hyperref}       
\usepackage{url}            
\usepackage{booktabs}       
\usepackage{amsfonts}       
\usepackage{nicefrac}       
\usepackage{microtype}      
\usepackage{lipsum}		
\usepackage{graphicx}
\usepackage{natbib}
\usepackage{doi}
\usepackage{multirow}
\usepackage{listings}
\usepackage{xcolor}
\usepackage{makecell}
\usepackage{xspace}
\usepackage{tcolorbox}

\newcommand{\pgcoder}{PanGu-Coder2\xspace}

\global\def\best{$62.20$}
\global\def\rrtf{RRTF\xspace}

\title{\pgcoder: Boosting Large Language Models for Code with Ranking Feedback }

\author{
Bo Shen\textsuperscript{*}~~Jiaxin Zhang\textsuperscript{*}~~Taihong Chen\textsuperscript{*}~~Daoguang Zan\textsuperscript{\S}~~Bing Geng\textsuperscript{*}~~An Fu\textsuperscript{*}~~Muhan Zeng\textsuperscript{*} \\ 
\textbf{Ailun Yu\textsuperscript{\dag}}~~\textbf{Jichuan Ji\textsuperscript{*}}~~\textbf{Jingyang Zhao\textsuperscript{*}}~~\textbf{Yuenan Guo\textsuperscript{*}}~~\textbf{Qianxiang Wang\textsuperscript{*}}
\\
\textsuperscript{*}Huawei Cloud Co., Ltd.\\
\textsuperscript{\S}Chinese Academy of Science \\
\textsuperscript{\dag}Peking University
}

\date{}

\renewcommand{\shorttitle}{\pgcoder: Boosting Large Language Models for Code with Ranking Feedback}

\hypersetup{
pdftitle={A template for the arxiv style},
pdfsubject={q-bio.NC, q-bio.QM},
pdfauthor={},
pdfkeywords={Large Language Model, Code Generation, Instruction Tuning, Reinforcement Learning},
}

\begin{document}
\maketitle

\begin{abstract}
Large Language Models for Code (Code LLM) are flourishing. New and powerful models are released on a weekly basis, demonstrating remarkable performance on the code generation task.
Various approaches have been proposed to boost the code generation performance of pre-trained Code LLMs, such as supervised fine-tuning, instruction tuning, reinforcement learning, etc.
In this paper, we propose a novel \textbf{RRTF} (\textbf{R}ank \textbf{R}esponses to align \textbf{T}est\&\textbf{T}eacher \textbf{F}eedback) framework, which can effectively and efficiently boost pre-trained large language models for code generation. Under this framework, we present \pgcoder, which achieves 62.20\% pass@1 on the OpenAI HumanEval benchmark.
Furthermore, through an extensive evaluation on CoderEval and LeetCode benchmarks, we show that \pgcoder consistently outperforms all previous Code LLMs.

\end{abstract}

\keywords{Large Language Model \and Code Generation \and Reinforcement Learning \and Instruction Tuning}

\section{Introduction}
As one of the most promising applications of large language model (LLM), code large language models have captivated considerable attention across academia and industry due to their remarkable capability in code-related tasks~\cite{zan2022neural}.
Since OpenAI released Codex~\cite{codex}, AlphaCode~\cite{alphacode}, PaLM-Coder~\cite{palm}, and PanGu-Coder~\cite{pangu-coder} are subsequently published but in a closed-source way. Researchers open-source CodeParrot~\cite{codeparrot}, PolyCoder~\cite{polycoder}, PyCodeGPT~\cite{cert}, and SantaCoder~\cite{santacoder}, but they fall far behind commercial models in terms of model size, capability, and performance.
The situation is changed by Hugging Face\footnote{\url{https://huggingface.co}}, as the BigCode community releases StarCoder~\cite{starcoder}: a 15B parameter model with 8K window size and FIM (Fill In the Middle, or infilling) capability. StarCoder outperforms many previous open-source large language models that support generating code from natural language descriptions, and even matches the OpenAI code-cushman-001 model on the HumanEval~\cite{codex} and MBPP benchmarks~\cite{mbpp}.

However, most large language models for code still fall behind the latest commercial models like GPT-3.5 and GPT-4 from OpenAI~\cite{gpt4-report, sparks}.
We use Code LLM to denote the large language model majorly pre-trained on code corpus, like PanGu-Coder~\cite{pangu-coder}, Replit~\footnote{\url{https://github.com/replit/ReplitLM}}, and StarCoder~\cite{starcoder}.
Compared with open-source Code LLMs, the OpenAI GPT-family models are usually bigger in size and majorly pre-train on natural language corpus (with a small proposition of code-related data), which can contribute to their superior natural language comprehension and instruction following capabilities.
Some efforts have been made to boost Code LLMs, like data engineering (phi-1~\cite{phi-1}), instruction tuning (WizardCoder~\cite{wizardcoder}), retrieval-augmented generation (ReAcc~\cite{reacc}, RepoCoder~\cite{repocoder}, etc.), and reinforcement learning (RLTF~\cite{rltf}, CodeRL~\cite{coderl}, PPOCoder~\cite{ppocoder}, etc.).

Although reinforcement learning (RL) seems to be a promising direction since programming is essentially a trial-and-error procedure, existing RL-based approaches face several major limitations.
The motivation is intuitive and straightforward: as we expect the model to generate code according to human intent and requirements, reinforcement learning on Code LLMs can help the model enhance the ability to interpret and respond to code generation instructions, thus increasing the likelihood of generating a code to successfully solve a given problem.
Typically, existing RL-based approaches design value/reward functions according to feedback signals from code processors, like compilers, debuggers, executors, and test cases. 
However, this leads to three limitations:
First, regarding the test results as a reward directly provides limited improvements to the base model.
Second, the adopted RL algorithm (like PPO) is complicated to implement and hard to train on large language models~\cite{rltf}.
Besides, running tests while training the model is time-consuming.
As a result, previous works~\cite{coderl,rltf} only experiment on modestly-sized models, and the improvement is rather limited.

To address the problem of existing RL-based approaches and further exploit the potential of Code LLM, we propose the \textbf{RRTF} (\textbf{R}ank \textbf{R}esponses to align \textbf{T}est\&\textbf{T}eacher \textbf{F}eedback) framework, which is a novel work to successfully apply natural language LLM alignment techniques on Code LLMs.
Different from previous works like CodeRL~\cite{coderl} and RLTF~\cite{rltf}, we follow the idea of RLHF (Reinforcement Learning from Human Feedback) that empowers InstructGPT/ChatGPT~\cite{instructgpt}, but implement a much simpler and efficient training approach using ranking responses as feedback instead of the absolute value of a reward model.

As a proof of concept, we apply RRTF on StarCoder 15B, and present a model that achieves the best performance among all published Code LLMs, namely the \pgcoder.
Through extensive evaluation on three benchmarks, including HumanEval, CoderEval, and LeetCode, we conjecture that Code LLMs do have the potential to surpass natural language models of the same or larger sizes on the code generation task. Furthermore, by analyzing the training process and manually inspecting the generation code samples, we highlight the importance of high-quality data in improving the models' instruction following and code writing capabilities.

In a nutshell, we make the following contributions:

\begin{itemize}
    \item We introduce a new optimization paradigm named \rrtf, which is a data-efficient, easy-to-implement, and model-agnostic framework to effectively boost the code generation performance of pre-trained Code LLMs.
    \item We present \pgcoder, a model that improves nearly 30\% over its base model and achieves new state-of-the-art performance on the HumanEval, CoderEval, and LeetCode benchmarks, surpassing all previously published Code LLMs.
    \item We share our experience and findings in constructing effective training data, training the model with RRTF, and optimizing such a model for fast inference.
\end{itemize}

\section{Related Work}
\label{sec.related}

\subsection{Large Language Model for Code (Code LLMs)}

As a momentous milestone, Codex~\cite{codex} boasting a $12$-billion-parameters model demonstrates the extraordinary capability to tackle up to $72\%$ of Python programming problems. Subsequently, a new wave of code generation models, such as AlphaCode~\cite{alphacode}, PaLM-Coder~\cite{palm}, and PanGu-Coder~\cite{pangu-coder}, also were proposed. Despite the remarkable prowess exhibited by the aforementioned models, it is disheartening to note their unavailability as open-source projects. Therefore, several open-source code generation models, including CodeParrot~\cite{codeparrot}, PolyCoder~\cite{polycoder}, PyCodeGPT~\cite{cert}, SantaCoder~\cite{santacoder}, and StarCoder~\cite{starcoder}, were released, injecting fresh vigor into the realm of code generation~\cite{codet}. Meanwhile, code generation models have also been applied to a broader range of practical coding scenarios. For example, CodeGeeX~\cite{codegeex}, BLOOM~\cite{bloom} and ERNIE-Code~\cite{ernie-code} have been proposed to facilitate multilingual modeling; JuPyT5~\cite{jupyt5} is trained on a large corpus of Jupyter notebooks, aiming to elevate the experience of interactive programming; DocCoder~\cite{doccoder} and APICoder~\cite{apicoder} have been proposed to empower language models with the ability to invoke APIs; Some models such as InCoder~\cite{incoder}, FIM~\cite{fim}, MIM~\cite{mim}, SantaCoder~\cite{santacoder}, and StarCoder~\cite{starcoder} support the code generation at arbitrary positions.

Of late, some efforts~\cite{lima,instruction-2} using the instruction tuning technique unlock the potential valuable knowledge stored within large language models, by fine-tuning on meticulously curated high-quality instruction datasets.
In the field of code generation, WizardCoder $15$B~\cite{wizardcoder} and phi-1 $1.3$B~\cite{phi-1} achieve exceptional code generation performance by fine-tuning on the data generated by OpenAI's GPT-3.5 or GPT-4.

\subsection{Reinforcement Learning on LLM}

\paragraph{Reinforcement Learning from Human Feedback}

Large language models can generate untruthful, unexpected, and unhelpful outputs, which are not aligned with the intention of the end users. To align the behavior of large language models with human intentions, \citet{DBLP:conf/nips/Ouyang0JAWMZASR22} proposed Reinforcement Learning from Human Feedback(RLHF) recently. The underlying idea is to leverage human preferences on given tasks to improve the behavior of a language model. A typical RLHF procedure consists of three steps, including supervised fine-tuning (SFT) which collects human demonstrations of desired model behavior and fine-tunes a language model, reward model (RM) training which employs humans to label the preferred output among various model outputs and trains a reward model based on the labeled data, and reinforcement learning via proximal policy optimization (PPO) which optimizes the language model against the reward model. OpenAI's GPT-3.5 and GPT-4 are trained with RLHF and their success demonstrates the effectiveness of RLHF to align the behavior of language models with human preferences. However, implementing RLHF requires heavy training resources and complex parameter tuning, which alleviates the technique from being easily applied in practice. In addition, the inefficiency and instability of RL algorithms can pose challenges to the alignment of language models. Given the limitations of heavy training resources and complex parameter tuning, \cite{rrhf} proposed the RRHF paradigm which leverages outputs with human preferences collected from various resources to train a model that aligns with human preferences. Its principle to align the model behavior to humans is to train a model to learn the outputs with better rewards based on human preferences among a set of outputs. Compared with RLHF, RRHF can be easily scaled to LLMs with larger sizes under a resource-constrained scenario. In view of the inefficiency and instability problem, \cite{dong2023raft} proposed the reward-ranked fine-tuning (RAFT) technique for language models. Their underlying idea is to first select high-quality outputs of the model based on the output ranking estimated by a reward model and then leverage the selected outputs to train a model that aligns with human preferences. Compared with RLHF, the SFT-style RAFT typically converges faster than the PPO used in RLHF, while utilizing simpler parameter configuration and fewer computational resources.

\paragraph{Reinforcement Learning on Code}

The successful practice of RLHF has inspired researchers to improve the capability of Code LLMs with reinforcement learning. For example, CodeRL~\cite{coderl} integrates actor-critic RL framework with unit test signals to fine-tune models. Following CodeRL, PPOCoder~\cite{ppocoder} uses the Proximal Policy Optimization (PPO) algorithm, but results in little improvements on the MBPP benchmark. Very recently, RLTF~\cite{rltf} moves a step forward by adopting an online RL framework with multi-granularity unit test feedback, to overcome the limitation of offline RL adopted by CodeRL and PPOCoder.

\subsection{Fine-tuning Code LLM}

Fine-tuning on pre-trained language models is a mainstream modeling paradigm that maximizes the performance at downstream tasks.
In the field of code, several works also adopt the paradigm to address code-related scenarios.
For instance, CodeGen~\cite{codegen} and StarCoder~\cite{starcoder} start by pre-training on a multilingual code corpus, followed by fine-tuning on monolingual data, thereby achieving superior performance on monolingual tasks.
Codex-S~\cite{codex} and PanGu-Coder-FT~\cite{pangu-coder} elevate their code generation capabilities by fine-tuning on competitive programming problems.
Recently, instruction tuning~\cite{instructgpt,gpt4-report}, as a form of supervised fine-tuning (SFT), is proposed to align the model with human behavior by learning abundant high-quality instruction corpus.
In this regard, WizardCoder~\cite{wizardcoder} was fine-tuned on a series of instruction corpora derived from a teacher model, effectively maximizing its code knowledge with relatively limited parameters.
In this technical report, \pgcoder employs ranking feedback strategy~\cite{rrhf} during the fine-tuning process, and achieves surprising code generation performance.

\section{Approach}
\label{sec.approach}

\subsection{Overview}

\begin{figure}[btp]
	\centering
        \includegraphics[width=0.8\linewidth]{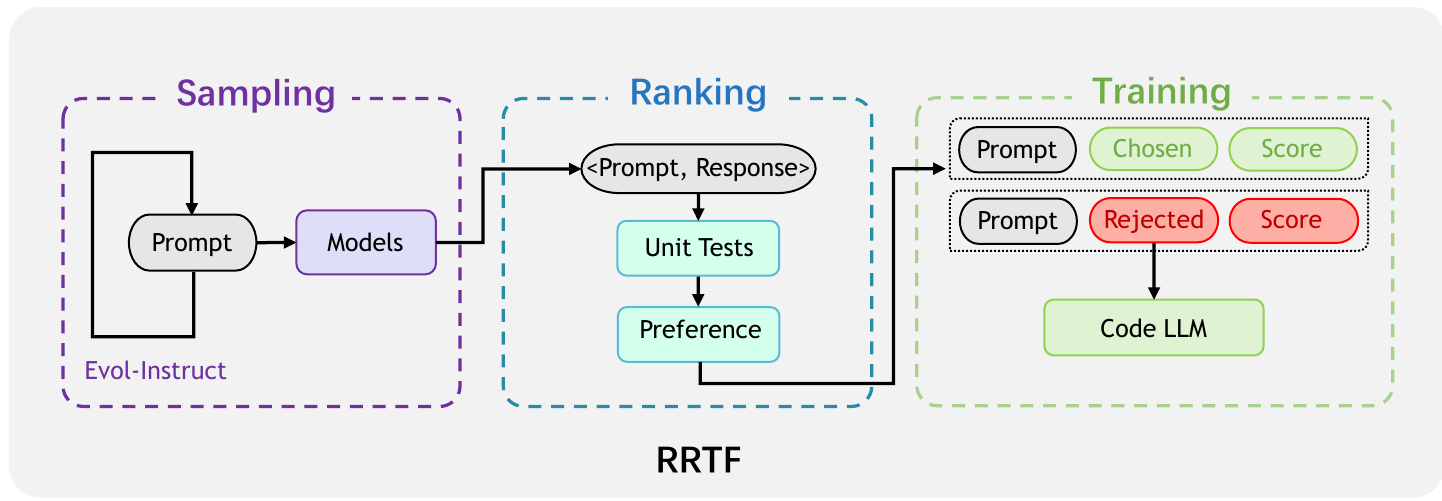}
	\caption{Overview of the proposed RRTF framework.}
	\label{fig.ov} 
\end{figure}

In this technical report, we present a simpler but powerful framework \textbf{RRTF}, which seamlessly combines several cutting-edge techniques, including instruct tuning~\cite{instruction-2}, Evol-Instruct method~\cite{wizardlm,wizardcoder}, and reinforcement learning~\cite{rrhf}.
The core idea of our approach is to guide a model towards producing higher-quality code, by utilizing the test signals and human preferences jointly as feedback to rank responses.
Inspired by recent progress in reinforcement learning and instruction fine-tuning on top of large natural language models, especially the RLHF~\cite{instructgpt}, RRHF~\cite{rrhf}, and RLTF~\cite{rltf}, we propose a new training paradigm, namely the RRTF framework.
Figure~\ref{fig.ov} shows the overview of the RRTF framework, which consists of three steps: sampling, ranking, and training. In the sampling stage, responses are sampled with prompts generated via Evol-Instruct. In the ranking stage, responses from different sources are ranked according to unit tests and heuristic preferences. In the training stage, triples of prompt and chosen/rejected responses with corresponding scores are used to train the Code LLM.

\subsection{Model Architecture}

In this work, we train a $15$B parameter model based on the decoder-only Transformer with Multi-Query-Attention\citet{DBLP:journals/corr/abs-1911-02150} and learned absolute positional embeddings.
At the same time, FlashAttention is used to reduce the amount of calculation and memory usage.
Hence, the max length of the model can be scaled to 8192.
Tabel \ref{tab.model_architecture} shows the detailed hyper-parameters of our model.

\begin{table}[h]
	\caption{The hyper-parameters of our model}
	\centering
	\begin{tabular}{lc}
		\toprule
		Hyper-Parameters  & \makecell{Value} \\
		\midrule
            Hidden size &  6144  \\
            Max Length 	& 8192 \\
            Num of attention heads 	& 48 \\
            Num of transformer hidden layers 	&  40  \\
		\bottomrule
	\end{tabular}
	\label{tab.model_architecture}
\end{table}

\subsection{Training Corpus}\label{training_corpus}

We follow the Evol-Instruct technique~\cite{wizardlm,wizardcoder} to construct our training corpus, since manually collecting a high-quality corpus is labor-intensive and time-consuming.
Specifically, we started from Alpaca 20K dataset\footnote{\url{https://huggingface.co/datasets/sahil2801/CodeAlpaca-20k}} and iteratively evolve the programming problems in this dataset via in-depth evolving to obtain new programming problems (the prompt is shown in Figure~\ref{fig:prompt}). With these problems, we sampled answers from different models. In total, we collected an initial corpus containing $100$K programming problems with answers, which we refer to as instruction and solution pairs. In addition, we conducted data preprocessing on our initial corpus using several manually-defined rules and reduced the size of the corpus to $68$K. 
More importantly, to prevent data leakage, we devoted considerable efforts to surveying the potential overlap between the collected $68$K dataset and the HumanEval benchmark. After conducting a meticulous survey, we confirm that there is no data leakage in our experiments, further validating the effectiveness of \pgcoder.

\begin{figure}
    \centering
\begin{tcolorbox}[left=2pt,right=2pt,top=0pt,bottom=0pt]
I want you to act as a Programming Contest Designer. Your objective is to rewrite a programming task based on the given task by increasing the difficulty a bit.\\
You can increase the difficulty using, but not limited to, the following methods:\\
\{methods\} \\

Your response is the rewritten programming task (\#Rewritten Task\#).\\
The \#Rewritten Task\# must be reasonable and must be understood and responded by humans, and also solvable with code. It should not be dependent on the \#Given Task\#. Your rewriting cannot omit the non-text parts such as the table and code in \#Given Task\#:. Also, please do not omit the input in \#Given Task\#.\\
**The rewritten task and the given task should have the similar length. **\\
**The rewritten task should ask for a function-level code solution.**\\
"\#Given Task\#", "\#Rewritten Task\#", "given task", and "rewritten task" are NOT allowed to appear in \#Rewritten Task\#. \\
\#Given Task\# \\
\{instruction\} \\
\#Rewritten Task\#
\end{tcolorbox}
\caption{Prompt to evolve over the CodeAlpaca dataset.}
\label{fig:prompt}
\end{figure}

\subsection{RRTF framework}

Inspired by RRHF~\cite{rrhf}, we propose the \textbf{RRTF} (\textbf{R}ank \textbf{R}esponses to align \textbf{T}est\&\textbf{T}eacher \textbf{F}eedback) framework for Code LLMs. RRHF~\footnote{\url{https://github.com/GanjinZero/RRHF}} is proposed as a simplified training paradigm for RLHF, which ranks responses from different sources according to human preferences, and aligns the model through a ranking loss function.
Compared with RLHF, RRHF can efficiently align the output probabilities of a language model with human preferences, with only 1-2 models required during the tuning period, and it is simpler than PPO in terms of implementation, hyperparameter tuning, and training.

Instead of aligning the model with human intents, the purpose of code generation is to improve generating correctness, so we replace the \textbf{H} (human) with \textbf{T}, which can be a combination of tests and teachers (more powerful models or human experts), they can jointly form a feedback signal to guide the generation of Code LLM and most of the feedback can be fully- or semi-automatically obtained in the faster way.
The training procedures of RRTF can be divided into 3 steps:

\begin{enumerate}
    \item \textbf{Step1: Sampling} In the sampling stage, responses are sampled with prompts. Based on the prompts generated by the Evol-Instruct (see Section~\ref{training_corpus}), we sample the responses both from the student model (model to train) and teacher models by various temperatures. The process is offline and in parallel, so we can efficiently get enough samples for training.
    \item \textbf{Step2: Ranking} In the ranking stage, responses from different sources are ranked according to unit tests and heuristic preferences. After obtaining all responses, we extract the programs from the responses and execute them in a running environment that supports large-scale parallel execution. According to the test results, there are 4 situations, which are \textit{compiled error}, \textit{runtime error}, \textit{pass partial tests}, \textit{all pass}. For each data, we assign different scores from low to high based on the above situations. Meanwhile, we filter out data whose teachers' score is lower than the student model. For two samples that fall into the same situation, we always assign a higher rank to the sample from the teachers, since we \textit{prefer} the student to learn from the teacher.
    \item \textbf{Step3: Training} In the training stage, triples of prompt and chosen/rejected responses with corresponding scores are used to train the Code LLM. During training, for each prompt $x$, we have a pair of response $\{y_{tea}, y_{stu}\}$, where $y_{tea}$ is the response generated by the teachers, and $y_{stu}$ is the response generated by the student model. So we can indicate the conditional log probability(length-normalized) $p_i$ by:
    $$
    p_i=\frac{\sum_t \log P_\pi\left(y_{i, t} \mid x, y_{i,<t}\right)}{\left\|y_i\right\|}
    $$
    where $\pi$ is the model, $i\in\{tea, stu\}$, $t$ is the time step. And the rank loss can be expressed as:
    $$
    L_{rank}={-\sum_{r_{tea}>r_{stu}}(r_{tea}-r_{stu})\min\left(0,p_{tea}-p_{stu}\right)}
    $$
    where $r_{tea}$ and $r_{stu}$ are the scores given in ranking stage. There is also a cross-entropy loss similar to supervised fine-tuning, which lets the model learn the response generated by the teacher:
    $$
    L_{f t}=-\sum_t \log P_\pi\left(y_{tea, t} \mid x, y_{tea,<t}\right)
    $$
    Finally, the total loss is the sum of the above two losses:
    $$
    L=L_{r a n k}+L_{f t}
    $$
\end{enumerate}

\subsection{Implementation Details}

We choose the StarCoder 15B~\cite{starcoder} as the base model, and train it with a global batch size of 512 for 6 epochs. 

\begin{figure*}[h]
  \centering
  \includegraphics[width=0.9\linewidth]{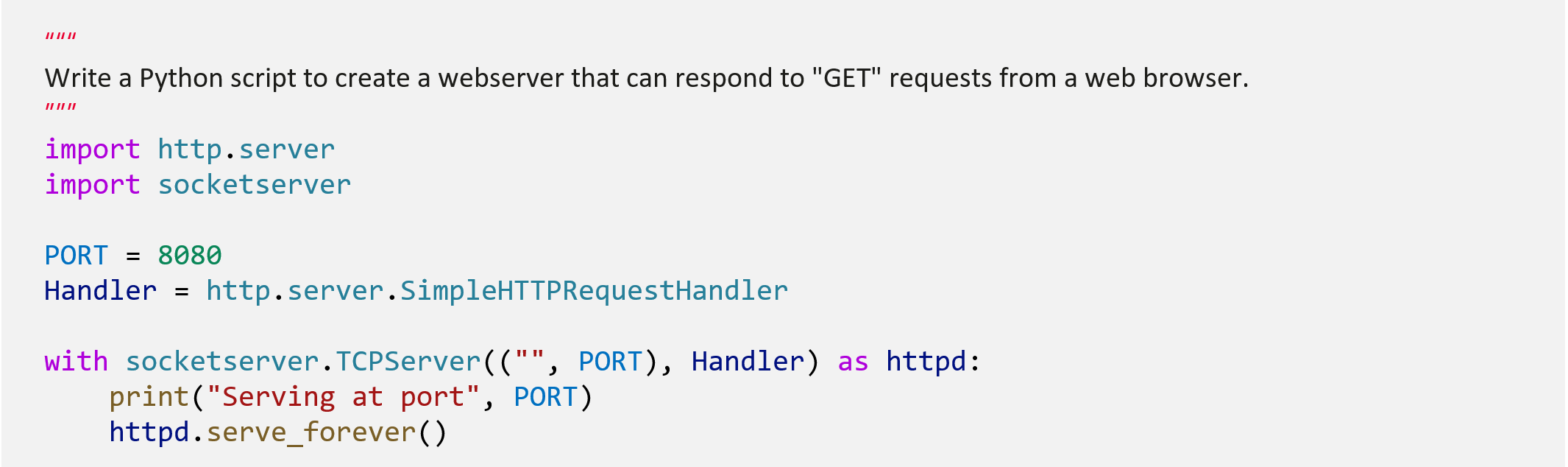}
  \caption{Example data format of the training sample.}
\label{fig.data_format}
\end{figure*}

Figure \ref{fig.data_format} shows the format of one single training sample. In addition to adding a pair of triple quotation marks on the prompt, we only use the code snippets extracted from responses for training.

\section{Evaluation}
\label{sec.exp}
We have conducted an extensive evaluation to study the performance of our approach. This section describes the settings of our evaluation and reports the experimental results as well as our findings.

\subsection{Evaluation Setup}

\subsubsection{Main Evaluated Models}

\begin{itemize}
    \item \textbf{CodeGen-mono 16B}~\cite{codegen} is a variant of CodeGen-Multi $16$B, specifically fine-tuned using additional Python code from GitHub.
    \item \textbf{CodeGeeX 13B}~\cite{codegeex} is a multilingual language model for code with a parameter count of $13$B, which is trained on approximately $850$B tokens from $23$ programming languages.
    \item \textbf{StarCoder 15B}~\cite{starcoder} is a Code LLM with $15$B parameters and a context size of $8$K, which supports infilling capabilities and fast inference.
    \item \textbf{CodeT5+ 16B}~\cite{codet5plus}, an encoder-decoder Code LLM, boasts modular flexibility, accommodating diverse code-related downstream tasks.
    \item \textbf{WizardCoder 15B}~\cite{wizardcoder} is the state-of-the-art Code LLM prior to \pgcoder, and is trained using the Evol-Instruct technique.
\end{itemize}

\subsubsection{Benchmarks}
\begin{itemize}
    \item HumanEval:\footnote{\url{https://github.com/openai/human-eval}} Released alongside Codex by OpenAI~\cite{codex}, the most widely-adopted benchmark for comparing LLMs' performance on code generation. HumanEval consists of 164 manually-written programming tasks.
    \item CoderEval~\cite{codereval}: A pragmatic code generation benchmark to evaluate models under realistic software development scenarios, including 230 functions with tests from 43 open-source Python projects.
    \item LeetCode (after 2022.7): We collected problems from leetcode that meet the following criteria:
    \begin{itemize}
        \item Problems that are publicly available and can be accessed for free.
        \item Problems that were created after July 1st, 2022, which ensures that any data in this benchmark does not overlap with the training data of StarCoder, which only consists of code before June 2022.
    \end{itemize}
        Besides the problem description, we also collected Python editor hints including method name and signature. We took editor hints as prompt input and tested models' output using public tests.
        As a result, this benchmark includes a total of 300 problems(with problem id $\geq$ 2325), including 79 easy problems, 150 medium problems, and 71 hard problems.
\end{itemize}

\subsubsection{Metric}

\paragraph{Pass@k} Same as related works, we also adopt the \textit{pass@k} metric implemented by OpenAI~\cite{codex} to assess the functional correctness of generated code, where $n (n \geq k)$ code samples are generated for each problem, and the number of correct samples $c$ is counted. The functional correctness of a code sample is determined by executing the corresponding unit tests and checking if it passes all test cases. Given the total number of generation $n$, the number of correct samples $c$, and the sampling budget $k$, \textit{pass@k} is calculated via the unbiased estimator:

$$
pass@k := \mathbb{E}[1-\frac{{n-c \choose k}}{{n \choose k}}], n=200, k \in \{1, 10, 100\}
$$

\subsubsection{Decoding Strategy} For experiments that evaluate the performance of models on code generation by estimating pass@k, we used a temperature of 0.2 to generate responses for pass@1, and a temperature of 1.2 for pass@10 and pass@100 for more diversity. For closed-source models, we retrieved the data from previous papers. For available models, we generated 200 samples to guarantee a statistically reliable result as much as possible. Additionally, we used a top\_p of 0.95 for nucleus sampling. For comparison of \pgcoder with other latest open-source models on three benchmarks, we used the greedy decoding strategy.

\subsubsection{Prompts} We noticed that the performance of a Code LLM could be largely affected by the prompt used for generating solutions to a programming problem. To maintain consistency with existing studies, for a given Code LLM, we leveraged the prompt reported in its corresponding paper to conduct our evaluation. The detailed code generation prompt for \pgcoder and other models are as follows:

\begin{tcolorbox}[left=2pt,right=2pt,top=0pt,bottom=0pt]
\textbf{Prompt for \pgcoder}\\
"""\\
\{docstring\}\\
"""\\
\{function signature\}\\
\\
\textbf{Prompt for StarCoder}\\
\{function signature\}

\hspace{1.5em}"""

\hspace{1.5em}\{docstring\}

\hspace{1.5em}"""\\

\textbf{Prompt for WizardCoder}\\
Below is an instruction that describes a task, paired with an input that provides further context. Write a response that appropriately completes the request. \\
\#\#\# Instruction:\\
Create a Python Script for this problem:\\
\{function signature\}

\hspace{1.5em}"""

\hspace{1.5em}\{docstring\}

\hspace{1.5em}"""\\

\#\#\# Response:
\end{tcolorbox}

\subsection{Evaluation Results}

\subsubsection{Performance}
We compared \pgcoder with existing Code LLMs in terms of Python code generation performance. 
Table~\ref{tab.pass1-10-100} shows the comparison result of pass@k on the HumanEval benchmark. Across all open-source models, \pgcoder achieves the best results for all $k$ values (pass@1=61.64, pass@10=79.55, pass@100=91.76). Compared with WizardCoder which was the state-of-the-art Code LLM on the HumanEval benchmark, we can observe that \pgcoder outperforms WizardCoder by a percentage of 4.34\%. 
With regard to StarCoder, we can observe 28\% absolute improvement in terms of pass@1 score (from 33.6\% to 61.6\%). In addition, for pass@10 and pass@100, the performance of \pgcoder is consistently better than that of StarCoder. 

Across all closed-source models, \pgcoder attains the second position. Compared with larger models including PaLM-Coder and LaMDA, \pgcoder performs better despite being smaller in scale. Another promising observation is that \pgcoder outperforms OpenAI's GPT-3.5. However, there is still a gap between our model and OpenAI's GPT-4 (the version reported in OpenAI's report~\cite{gpt4-report}).

Table~\ref{tab.pass1-greedy-decoding} shows the comparison result of greedy decoding pass@1. Across all benchmarks, we can observe that \pgcoder achieves the best results among all models, with a pass@1 value of 62.20\% on HumanEval, 38.26\% on CoderEval, and 32/30/10 on LeetCode. A promising observation is that \pgcoder not only surpasses WizardCoder and StarCoder on HumanEval, but also outperforms these two models on CoderEval and LeetCode. This indicates that \pgcoder not only excels at simple programming tasks, but also performs outstandingly well on context-aware development tasks and programming contest problems. 

From the experimental results shown in Tables~\ref{tab.pass1-10-100} and \ref{tab.pass1-greedy-decoding}, we can conclude that:
\begin{itemize}
    \item \pgcoder achieves a state-of-the-art 61.64\% pass@1 on HumanEval among open-source models.
    \item \pgcoder outperforms models of larger scale including PaLM-Coder and LaMDA despite being smaller in scale.
    \item \pgcoder is the only model we tested that achieves the best performance on HumanEval, CoderEval, and LeetCode at the same time.
\end{itemize}

\begin{table}
	\caption{Results of pass@1/10/100 of well-known models on HumanEval. Most scores are retrieved from previous papers as they are reported. For \pgcoder, we follow the Codex~\cite{codex} and AlphaCode~\cite{alphacode} paper to generate n=200 samples and report the optimal pass@1/10/100 when temperature=0.2/1.2/1.2 and top\_p=0.95. The same settings are used for StarCoder and WizardCoder (marked with *).}
	\centering
	\begin{tabular}{lrlll}
		\toprule
		\multirow{2.5}{*}{Model}  & \multirow{2.5}{*}{Params} & \multicolumn{3}{c}{Pass@k (\%)} \\ 
            \cmidrule(r){3-5} 
            & & k=1 & k=10 & k=100 \\
            \midrule
            \multicolumn{5}{c}{Closed-source Models} \\
		\midrule
            AlphaCode~\cite{alphacode}  & 1.1B & 17.1 & 28.2 & 45.3 \\
            Phi-1~\cite{phi-1} & 1.3B & 50.6 & - & - \\
            Codex~\cite{codex} & 12B & 28.81 & 46.81 & 72.31 \\
            LaMDA~\cite{lamda} & 137B & 14.0 & - & 47.3 \\
            PaLM-Coder~\cite{palm} & 540B & 36.0 & - & 88.4 \\
            GPT-3.5~\cite{gpt4-report} & - & 48.1 & - & - \\
            GPT-3.5~\cite{wizardcoder} & - & 68.9 & - & - \\
            GPT-4~\cite{gpt4-report} & - &  67.0 & - & - \\
            GPT-4~\cite{sparks} & - &  82.0 & - & - \\
            \midrule
            \multicolumn{5}{c}{Open-source Models} \\
            \midrule
            CodeGen-mono~\cite{codegen} & 16B & 29.28 & 49.86 & 75.00 \\
            CodeGeeX~\cite{codegeex} & 13B & 22.89 & 39.57 & 60.92 \\
            StarCoder~\cite{starcoder}\textsuperscript{*}   & 15B & 33.60 & 45.78 &  79.82  \\
            CodeT5+~\cite{codet5plus} & 16B & 30.9& 51.6& 76.7\\
            WizardCoder~\cite{wizardcoder}\textsuperscript{*}  & 15B  & 57.30 & 73.32 & 90.46  \\
            \midrule
            \textbf{\pgcoder}\textsuperscript{*} & 15B & \textbf{61.64} & \textbf{79.55} & \textbf{91.76}\\
		\bottomrule
	\end{tabular}
	\label{tab.pass1-10-100}
\end{table}

\begin{table}
	\caption{Performance comparison of \textit{\pgcoder} with previous models on three benchmarks by greedy decoding.}
	\centering
	\begin{tabular}{lrcccc}
		\toprule
		Model  & Params & \makecell{HumanEval \\ (text2code)}     & \makecell{CoderEval\\ (context2code)} & \makecell{LeetCode\\ (easy/medium/hard)} \\
		\midrule
            PanGu-Coder 	& 2.6B & 23.78 & 15.21  &  6/3/0  \\
            Replit-code-instruct-glaive\footnote{\url{https://huggingface.co/sahil2801/replit-code-instruct-glaive}} & 2.7B & 56.10 & 27.39  & 3/5/2\\
		StarCoder  & 15B  & 32.93 & 37.82  & 18/13/2       \\
  		WizardCoder  & 15B & 59.80 & 33.48  & 29/22/7      \\
            \textbf{\pgcoder} & 15B & \textbf{62.20} & \textbf{38.26} & \textbf{32/30/10}\\
		\bottomrule
	\end{tabular}
	\label{tab.pass1-greedy-decoding}
\end{table}

\subsubsection{Findings}
To analyze the training process of \pgcoder, we focus on two of the key factors that affect the performance of large language models: the dataset size and the training compute.

\begin{figure*}[tp]
  \centering
  \includegraphics[width=1.0\linewidth]{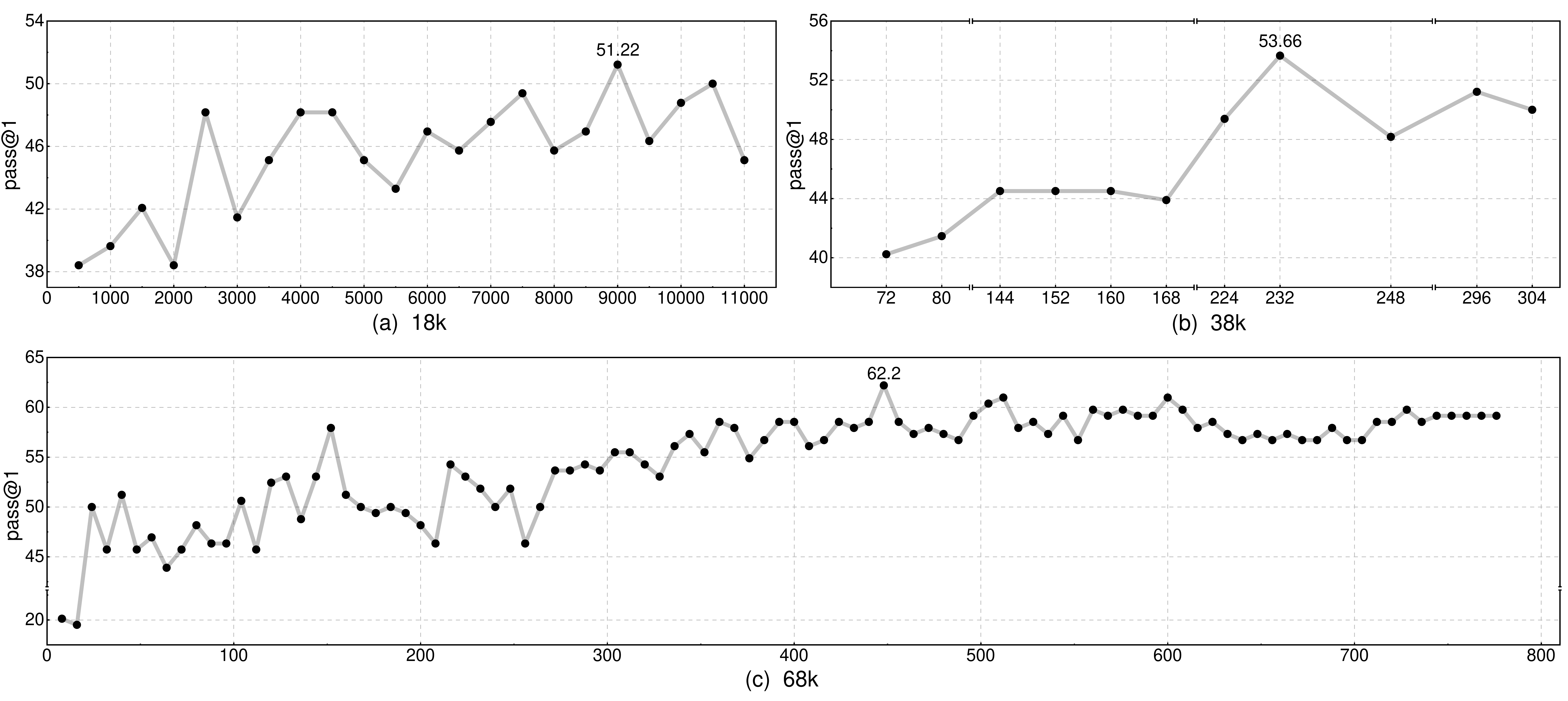}
  \caption{Performance change when in the training process (pass@1 on HumanEval with greedy decoding). The number of steps in an epoch for (a),(b), and (c) is roughly 2250, 74, and 132 respectively.}
\label{fig.training_curve}
\end{figure*}

\paragraph{Dataset size}
The overall accuracy (estimated via greedy decoding pass@1) increases along with the growth of dataset size, as shown in Figure \ref{fig.training_curve}.
Also, as the size of the dataset grows, the training curve becomes more stable, at roughly 2-3 epochs on 38k/68k dataset. As for the 18k dataset, performance still oscillates drastically after 3 epochs.
This suggests that more and variant corpus can result in better performance, while the training cost is still acceptable as epochs needed for reaching the best performance do not increase along with the scale of the corpus. 

\paragraph{Training compute}

Regardless of dataset size, the accuracy may drop drastically or stay flat at the start of the training.
After roughly 2 epochs, the training curve becomes more stable and the accuracy consistently increases as the loss decreases. 
The best performances are reached after 3 epochs while the accuracy becomes even more stable after 4 epochs, showing a sign of convergence.
This suggests that the model needs roughly 3-4 epochs to fully capture the knowledge in the dataset, and training steps after that may have very little help towards increasing the model's capability.

\subsubsection{Case Study}

To empirically study the model and shed light on future work, we compare and analyze the successful and failed cases of three models: the base model StarCoder, the instruction-tuned model WizardCoder, and the \pgcoder model.

\begin{figure*}[tp]
  \centering
  \includegraphics[width=0.8\linewidth]{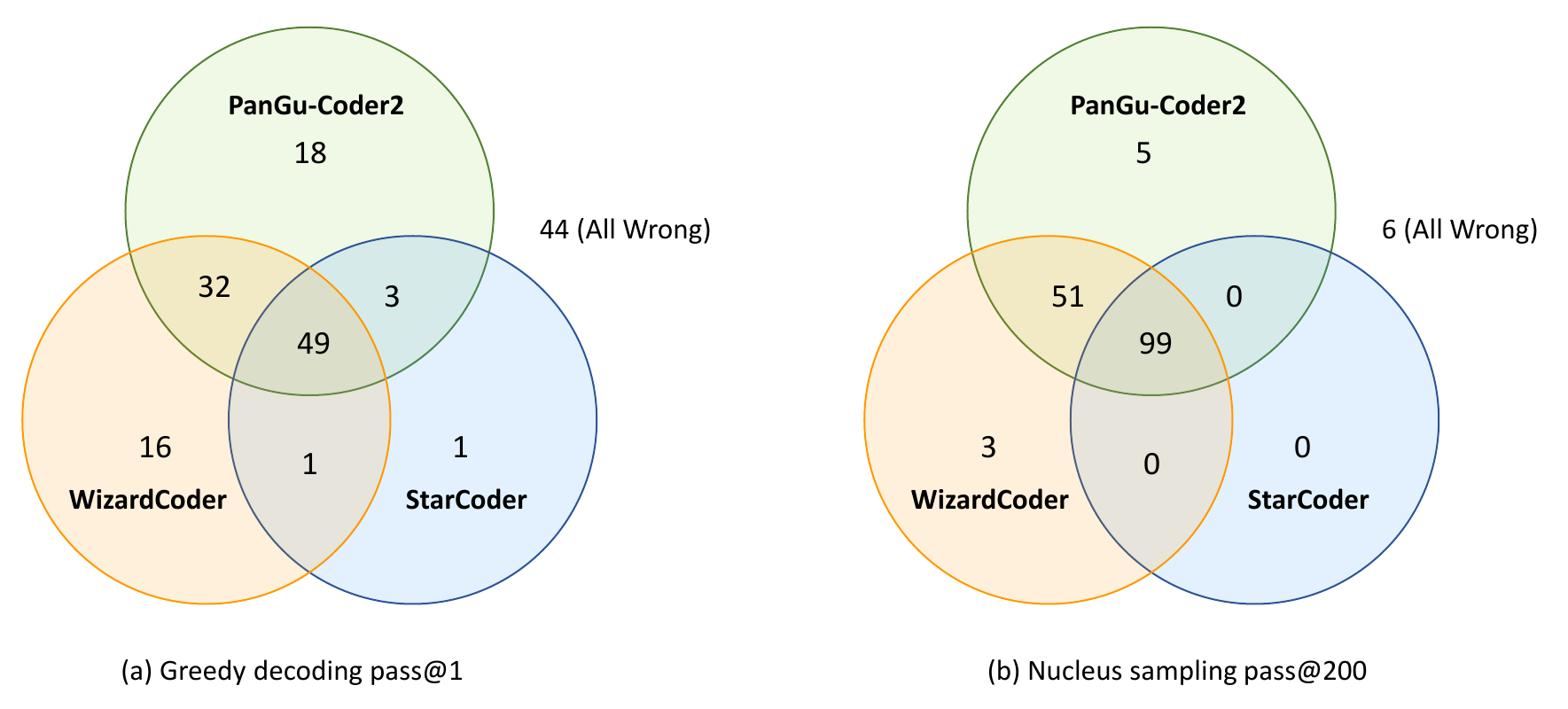}
  \caption{Numbers of correctly-solved problems by three models on HumanEval. }
\label{fig.overlap}
\end{figure*}

Figure~\ref{fig.overlap} shows the difference and intersection of solved problems by three models, in terms of greedy decoding and nucleus sampling. From the figure, we find that \pgcoder and WizardCoder can be complementary: though \pgcoder solves the most problems and some of them cannot be solved by WizardCoder, there are problems that were only solved by WizardCoder, which boosts the StarCoder performance in the instruct tuning way. Besides, there are some problems that cannot be solved by any of these models, even sampling for 200 times.

\begin{figure*}[hbp]
  \centering
  \includegraphics[width=1\linewidth]{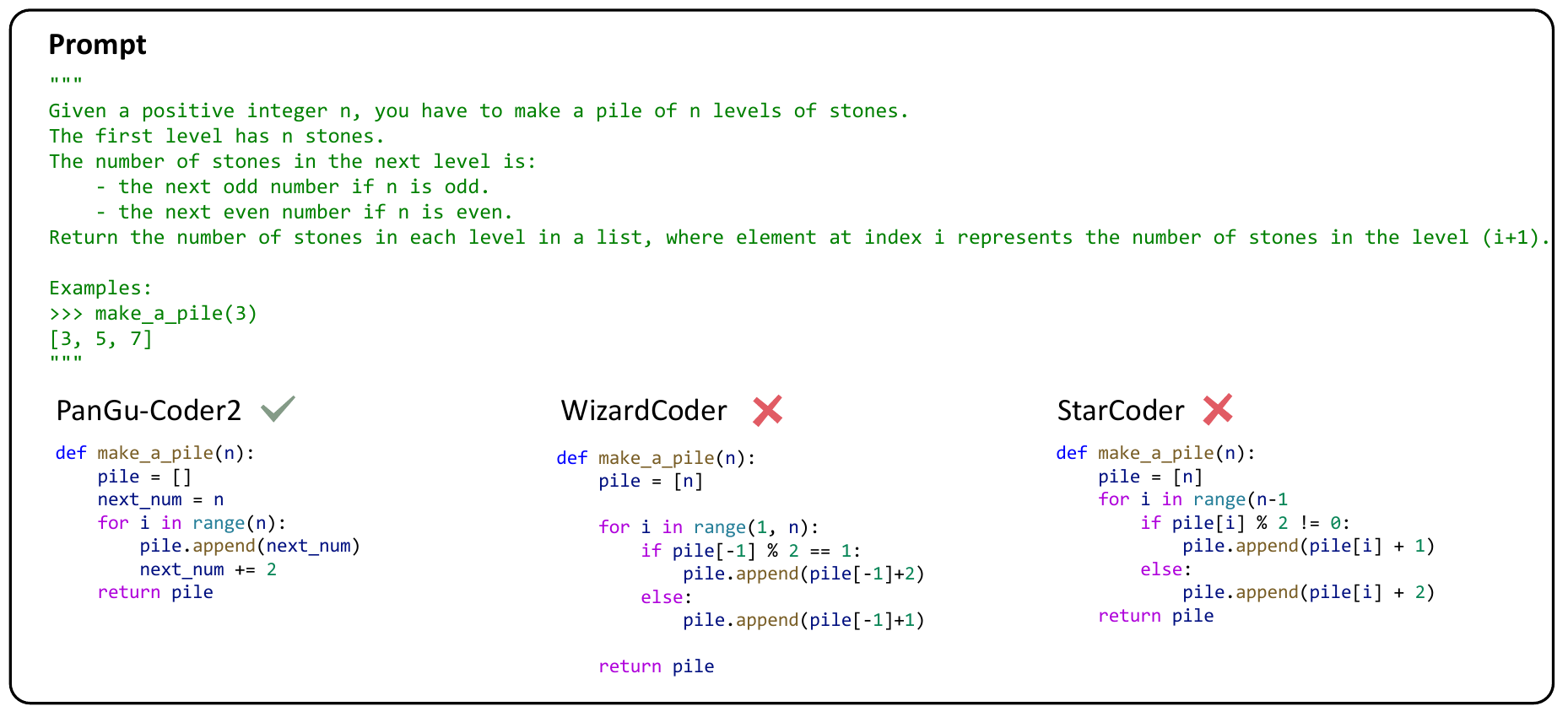}
  \caption{A HumanEval case of \pgcoder surpassing WizardCoder and StarCoder.}
\label{fig:pangu_right}
\end{figure*}

\begin{figure*}[tp]
  \centering
  \includegraphics[width=1\linewidth]{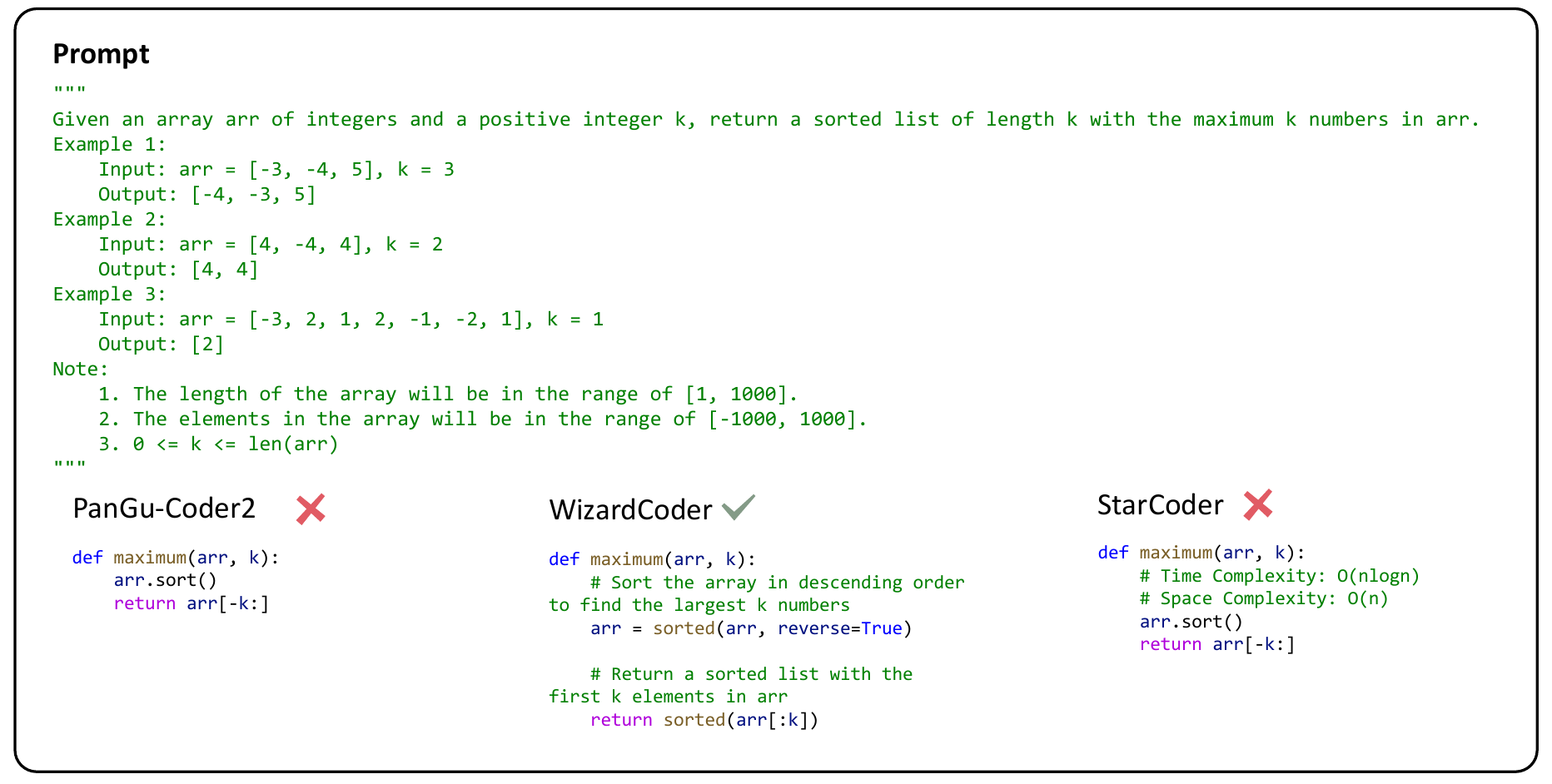}
  \caption{A HumanEval case of WizardCoder surpassing \pgcoder and StarCoder.}
\label{fig:wizard_right}
\end{figure*}

\begin{figure*}[tp]
  \centering
  \includegraphics[width=1\linewidth]{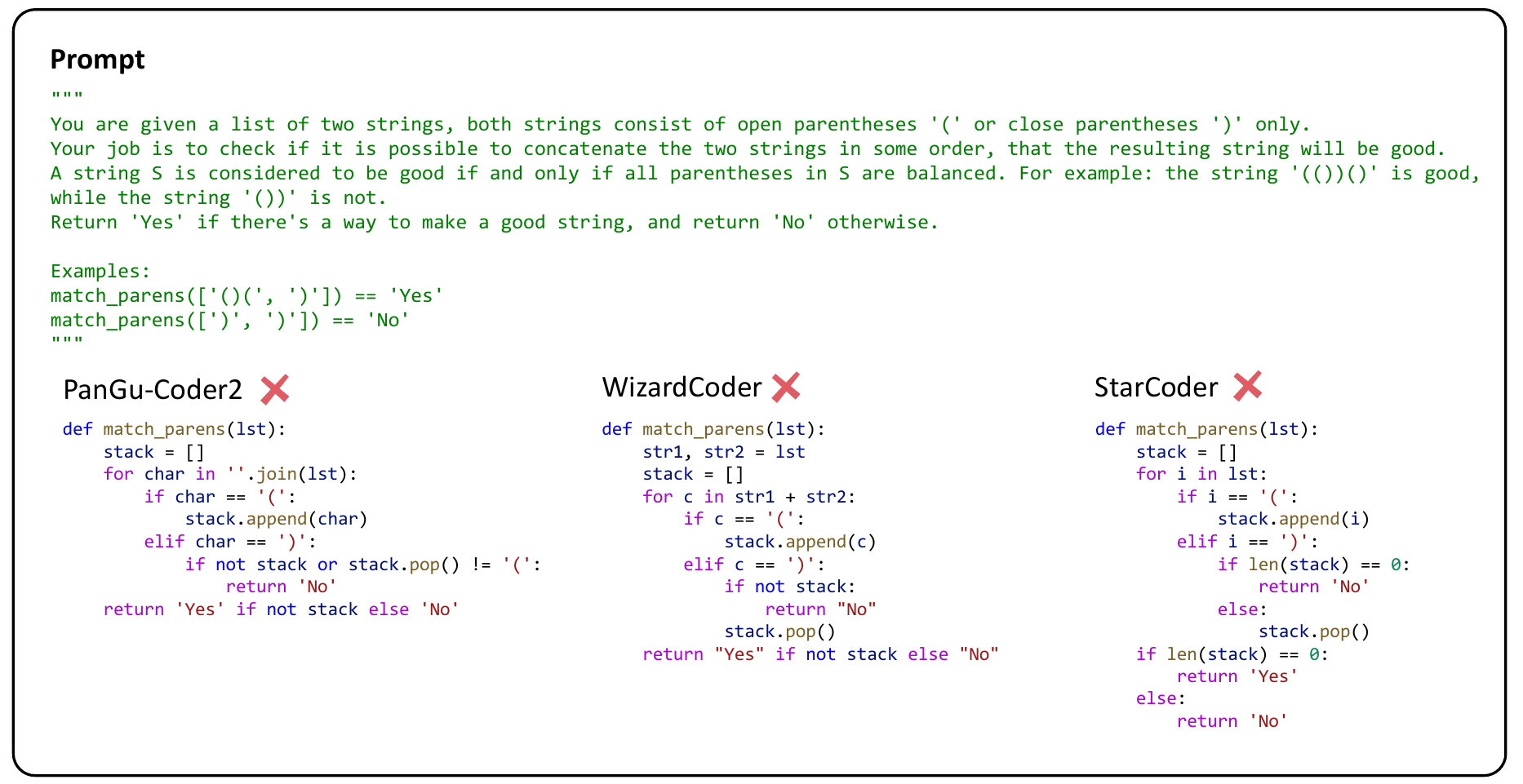}
  \caption{A HumanEval case where \pgcoder, WizardCoder, and StarCoder all generate incorrect outputs.}
\label{fig:non_right}
\end{figure*}

We choose several representative example codes generated by StarCoder, WizardCoder, and \pgcoder as the case study, to conduct a critical evaluation of the strengths and weaknesses of \pgcoder.
As depicted in Figure~\ref{fig:pangu_right}, \pgcoder adeptly comprehends the logical aspects of programming problems, while WizardCoder and StarCoder fall short in this regard.
This observation signifies that \pgcoder has effectively established an meticulous mapping between programming problem statements and code solutions via our proposed ranking feedback.
According to the depiction in Figure~\ref{fig:wizard_right}, in certain instances, \pgcoder and StarCoder are outperformed by WizardCoder, which may benefit from training with extensive rich-comment code instructions.
To a certain extent, this observation implies that incorporating step-by-step comment data may yield positive effects during the training process.
In addition, 
Figure~\ref{fig:non_right} shows a case where StarCoder, WizardCoder, and \pgcoder all give incorrect code solutions due to the intricacies and challenges posed by the programming problem.
This observation indicates that currently code LLMs still fall short of human-like proficiency in tackling complex programming requirements, leaving some room for improvement.

\subsection{Inference Optimization}
Since GPU memory consumption and inference speed are crucial factors for the deployment and use of model in practice, we conducted experiments involving the following quantization techniques to study the optimization strategies of model inference:
\begin{itemize}
    \item CTranslate2:\footnote{\url{https://github.com/OpenNMT/CTranslate2}} CTranslate2 is a library for accelerating Transformer models when inference, developed by OpenNMT.
    \item GPTQ:\footnote{\url{https://github.com/PanQiWei/AutoGPTQ}} A LLMs quantization package based on GPTQ algorithm.
\end{itemize}
Table~\ref{tab.inf} shows the GPU memory consumption, inference speed, and HumanEval performance of models optimized using different quantization techniques. We used 8-bit (4-bit) quantization and the following decoding parameters in the inference stage: top\_p=0.95, tempreture=0.2, max\_new\_tokens=512. Across all quantization techniques, we can observe a significant decrease in terms of memory usage and a significant increase in terms of inference speed. It is incredible that after being quantized with CTranslate2, the performance of our model on HumanEval has a slight improvement. A plausible reason for this phenomenon is the robustness of \pgcoder itself. We plan to conduct an in-depth study on this interesting result in our further work.

\begin{table}[htp]
	\caption{A comparison of different quantization techniques (on the same device)}
	\centering
	\begin{tabular}{lcccc}
		\toprule
		Model  & Precision & \makecell{GPU Memory Consumption\\ (GB)} & \makecell{Inference Speed \\(ms/token)} & \makecell{HumanEval \\(greedy decoding)} \\
		\midrule
            \pgcoder & float16 &32.36  &75   & \best \\
            \pgcoder-CTranslate2 & int8 &16.29  &33  &64.63  \\
            \pgcoder-GPTQ  & int8 &16.92  & 51  &51.22  \\
            \pgcoder-GPTQ  & int4 &9.82  & 42  &51.83  \\
		\bottomrule
	\end{tabular}
	\label{tab.inf}
\end{table}

\section{Conclusion}
In this paper, we introduce a novel frameork, namely \rrtf, and present a new Code LLM, namely \pgcoder.
Firstly, we adopt the Evol-Instruct technique to obtain a substantial amount of high-quality natural language instruction and code solution data pairs.
Then, we train the base model by ranking candidate code solutions using feedback from test cases and heurstic preferences. Through comprehensive evaluations on HumanEval, CodeEval, and LeetCode benchmarks, \pgcoder achieves new state-of-the-art performance among billion-parameter-level Code LLMs, surpassing all of the existing ones by a large margin. In our future work, we will delve into the combination of \rrtf and instruct tuning to boost the performance of Code LLMs.


\bibliographystyle{unsrtnat}
\bibliography{references}  






\end{document}